\documentclass[10pt,twocolumn,letterpaper]{article}

\usepackage{cvpr}
\usepackage{times}
\usepackage{epsfig}
\usepackage{graphicx}
\usepackage{amsmath}
\usepackage{amssymb}

\usepackage{bm}
\usepackage{amsfonts}
\usepackage{multirow}
\usepackage{algorithm}
\usepackage{algorithmic}
\usepackage{subfigure}
\usepackage{xcolor}
\usepackage{array}
\renewcommand{\multirowsetup}{\centering}

\def\eg{\emph{e.g.}}

\def\ie{\emph{i.e.}}

\usepackage[breaklinks=true,bookmarks=false]{hyperref}

\cvprfinalcopy % *** Uncomment this line for the final submission

\def\cvprPaperID{****} % *** Enter the CVPR Paper ID here

\begin{document}
%%%%%%%%% TITLE
\title{Unsupervised Person Re-identification via Softened Similarity Learning}

\author{Yutian Lin\textsuperscript{1}\thanks{This work was done when the first author was an intern at Huawei Noah's Ark Lab.},\quad Lingxi Xie\textsuperscript{2},\quad Yu Wu\textsuperscript{3,4},\quad Chenggang Yan\textsuperscript{1},\quad Qi Tian\textsuperscript{2}\thanks{Qi Tian is the corresponding author.}\\
\textsuperscript{1}Hangzhou Dianzi University,\quad  \textsuperscript{2}Huawei Inc., \quad
\textsuperscript{3}Baidu Research,\\ \textsuperscript{4}ReLER, University of Technology Sydney\\ {\tt\small yutianlin477@gmail.com},\quad{\tt\small 198808xc@gmail.com},\quad{\tt\small yu.wu-3@student.uts.edu.au},\\
{\tt\small cgyan@hdu.edu.cn},\quad{\tt\small tian.qi1@huawei.com}
}

\maketitle
\thispagestyle{empty}

\begin{abstract}
Person re-identification (re-ID) is an important topic in computer vision. This paper studies the unsupervised setting of re-ID, which does not require any labeled information and thus is freely deployed to new scenarios. There are very few studies under this setting, and one of the best approach till now used iterative clustering and classification, so that unlabeled images are clustered into pseudo classes for a classifier to get trained, and the updated features are used for clustering and so on. This approach suffers two problems, namely, the difficulty of determining the number of clusters, and the hard quantization loss in clustering. In this paper, we follow the iterative training mechanism but discard clustering, since it incurs loss from hard quantization, yet its only product, image-level similarity, can be easily replaced by pairwise computation and a softened classification task. With these improvements, our approach becomes more elegant and is more robust to hyper-parameter changes. Experiments on two image-based and video-based datasets demonstrate state-of-the-art performance under the unsupervised re-ID setting.
\end{abstract}

\section{Introduction}
Given a query image, person re-identification (re-ID) aims to match the person across multiple non-overlapped cameras. 
In the last few years, person re-ID has drawn increasing research attention \cite{li2014deepreid,zheng2015scalable,zheng2019joint,Tay_2019_CVPR,sun2018beyond,sun2019perceive}, due to its wide range of applications such as finding people of interest (\eg, lost kids or criminals) and person tracking. However, most of the proposed methods are of supervised manner, which requires intensive manual labeling and is not applicable to real-world applications. To relieve the scalability problem, in this paper, we focus on the unsupervised re-ID task.

\begin{figure}[!t]
\begin{center}
\includegraphics[width=\linewidth]{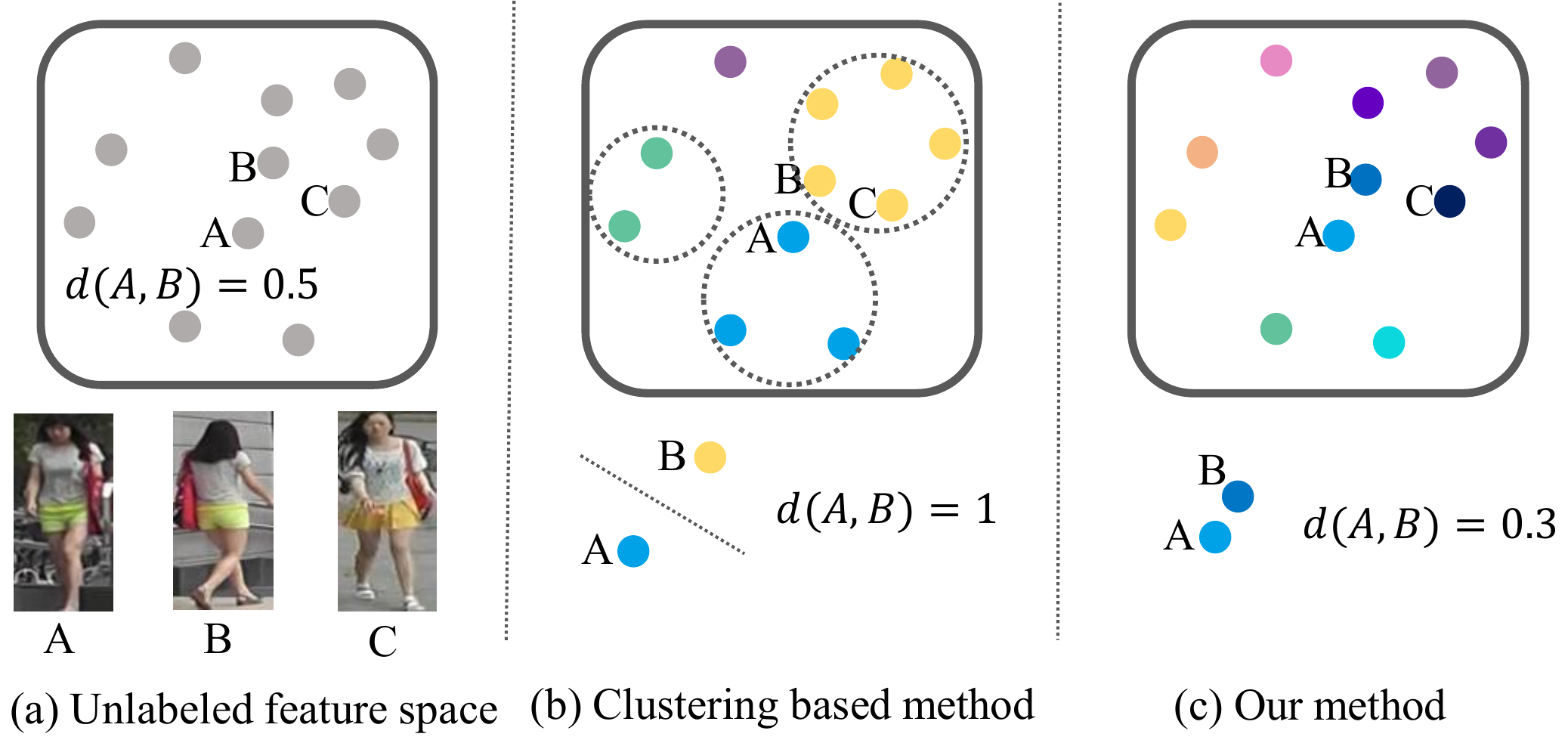}
\end{center}
\caption{(a) Unlabeled images are represented as gray circles in the feature space. The image A and B are of the same person, with an initialized distance of 0.5. The image C is from another person. (b) The clustering based unsupervised re-ID method roughly divides images into classes for network training. Although images A and B are of the same identity, they are assigned with different pseudo labels and learn to be separated. (c) Our method push circles in similar colors (similar images) closer with a soft constraint.
}
\label{fig:motivation}
\end{figure}

Different from the unsupervised domain adaptation (UDA) methods 
\cite{Wang_2018_CVPR,yang2019patch,yu2019unsupervised} that leverage the prior knowledge learned from other re-ID datasets, in this paper, we aim to solve the problem without any re-ID annotation.
A branch of methods \cite{fan18unsupervisedreid,lin2019aBottom,lin2020_unsupervised} were verified effective, which adopted an iterative clustering and deep learning mechanism, where the network was trained based on the pseudo labels generated by unsupervised clustering. 
However, the clustering based methods roughly divided images into clusters for training, which made the model highly depends on the clustering result. 
As shown in Figure \ref{fig:motivation} (b), images of the same person could be divided into different clusters, which are further trained to be separated with the wrong assigned pseudo label.
Since mistakes of unsupervised clustering are inevitable, learning with a hard quantization loss can be prone to fit the noisy labels produced by clustering.

In this paper, we propose a new framework of unsupervised learning in which clustering is no longer required, and thus the error of the hard quantization loss is relieved. As illustrated in Figure~\ref{fig:motivation} (c), instead of using explicit labels generated by clustering, we mine the relationship between unlabeled images as a gentle constraint to make similar images have closer representations.
Specifically, our framework adopts a classification network with softened labels, where the softened labels reflect the image similarity. Unlike the original one-hot labels that force images belonging to an exact class, we treat the labels as a distribution, that an image is encouraged to be associated with several related classes. For each training data, the network is trained not only to predict the ground-truth class, but motivated to predict the similar classes. The learned embedding is then close to similar ones and has a long distance from irrelevant images. 
On the one hand, without learning with the hard labels, the hard quantization error is eliminated. On the other hand, the supervision of the softened label is relatively weak, which also provides more room for the algorithm. 
In order to fully exploit the potential of the model, we introduce some auxiliary information to help find similar images. 
Specifically, when measuring the similarity between images, camera ID and partial details of each pedestrian image are studied. To relieve the issue of camera variance, we propose the cross-camera encouragement term (CCE) that promotes the softened similarity learning from images under different camera views. In this way, the model will learn from more diverse data. Note that the camera ID is automatically obtained at the moment of capturing and is no need for human labeling.
Moreover, we extract part features and consider the partial details along with the global appearance as an additional clue.

We evaluate the proposed method on two image-based and two video-based re-ID datasets. The experimental results reveal that our method is robust and stable during iterations via softened similarity learning. Our method outperforms state-of-the-art unsupervised methods on all the four datasets. With the high accuracy and the advantage that does not require any annotations, our approach is easy to be deployed in real-world applications.

Our contributions can be summarized in two-fold. \textbf{First}, we propose an unsupervised re-ID framework via softened similarity learning. A classification network is adopted with re-assigned soft label distribution to learn from similar images with smooth constraint. By pushing each person images to get closer to similar images and pushing all other person images away from each other, our framework learns a robust and discriminative model with high potential. \textbf{Second}, to make use of the high potential model, we introduce auxiliary information to guide similarity estimation. A cross-camera encouragement (CCE) term is proposed to encourage the similarity exploration between images of different camera views. The fine-grained details are also considered when measuring similarity. These strategies are also proven effective when plugging in other unsupervised re-ID methods. 

\begin{figure*}[!t]
\begin{center}
\includegraphics[width=\linewidth]{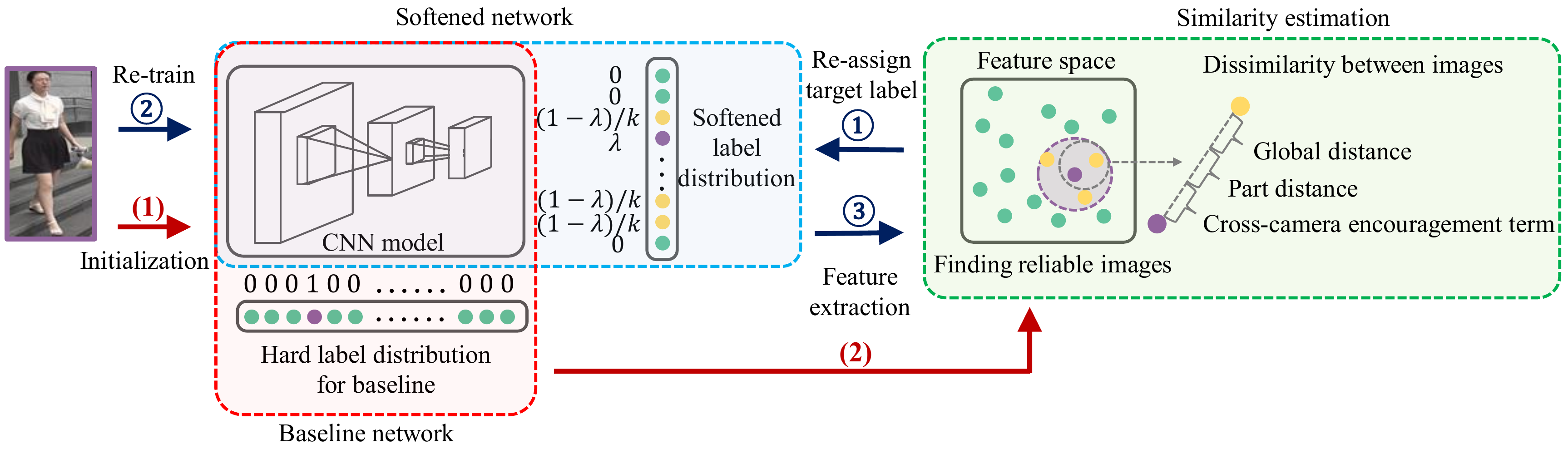}
\end{center}
\caption{Overview of our method. First, a baseline network with hard label distribution is adopted for initialization, which is shown following the red arrows. 
Subsequently, with the initialized network, three procedures are conducted iteratively: 1. Feature embeddings of training images are extracted; 2. Similarity among images is estimated to re-assign the target label; 3. The network is re-trained with the softened labels. 
These procedures are shown following the blue arrows. Notably, the procedures with red arrows are conducted once, while the procedures with blue arrows are conducted iteratively.
}
\label{fig:pipline}
\end{figure*}

\section{Related Works}
\subsection{Supervised Person Re-identification}
Most re-ID methods are in a supervised manner, in which sufficient labeled images are given. 
Recently, with the developing of deep learning approaches \cite{yan2019stat,yan20203d,yan2020deep}, methods with convolutional neural networks have dominated the re-ID community \cite{li2014deepreid,varior2016gated,zheng2015scalable,zheng2019joint,Tay_2019_CVPR,lin2019improving}.  
Specifically, methods proposed to learn discriminative features from parts of pedestrian images achieve impressive performance \cite{sun2018beyond,kalayeh2018human,sun2019perceive}. 
For example, in \cite{sun2018beyond}, the feature maps are cut into uniform pieces for classification, and the part-informed features are assembled as the descriptor. A refined part pooling is further proposed to reinforce the within-part consistency in each part.

In our paper, we focus on unsupervised re-ID without annotated labels. We take advantage of the strategy of learning from the part. To exploit the fine-grained information, we directly divide the global feature into horizontal pieces to measure the similarity between each pair of the corresponding parts.

\subsection{Unsupervised Domain Adaptation}
To relieve the scalability problem of supervised re-ID, some unsupervised domain adaptation methods (UDA) \cite{peng2016unsupervised,fan18unsupervisedreid,Wang_2018_CVPR,deng2018image,wei2018person,bak2018domain,liu2019adaptive} were proposed to learn a re-ID model from a labeled source domain and an unlabeled target domain.
Wang~\emph{et al.}~\cite{Wang_2018_CVPR} proposed to learn an attribute-semantic and identity discriminative representation from the source dataset, which is transferable to the target domain.
In \cite{yang2019patch}, a PatchNet pre-trained on the source dataset is used to generate pedestrian patches. A network is then designed to pull similar patches together and push the dissimilar patches.
In \cite{yu2019unsupervised}, a soft multilabel is learned for each unlabeled person by comparing the unlabeled person with a set of known reference persons from the source domain.
Zhong~\emph{et al.}~\cite{zhong2019invariance} proposed a framework that consists of a classification module and an exemplar memory module, which calculates the cross-entropy loss for labeled source data and saves the up-to-date features for target data and computes the invariance learning loss for unlabeled target data, respectively.

Unsupervised domain adaptation methods usually obtain impressive performance. However, these methods take advantage of the external source domain, which is annotated with cross-camera identity labels. In contrast, we focus on the fully unsupervised re-ID task without any external dataset or identity annotation.

\subsection{Unsupervised Person Re-identification}
The traditional unsupervised methods usually fall into three categories, designing hand-craft features \cite{gray2008viewpoint,farenzena2010person,liao2015person,lisanti2015person,matsukawa2016hierarchical}, exploiting localized salience statistics \cite{zhao2013unsupervised,zhao2013person,wang2014unsupervised} or dictionary learning based methods \cite{kodirov2015dictionary,kodirov2016person}. The performances of these methods are usually low, because it is challenging to design features for images captured by different cameras, under different illumination and view condition.
In \cite{yu2017cross}, camera information is used to learn view-specific projection for each camera view by jointly learning the asymmetric metric and seeking optimal cluster separations. However, this method is not suitable for dataset captured by multiple cameras, because the view-specific projection is learned from a pair of cameras.

Recently, Lin \emph{et al.} \cite{lin2019aBottom} proposed a bottom-up clustering framework that iteratively trains a network based on the pseudo label generated by unsupervised clustering. However, due to the clustering error, images could be assigned wrong pseudo labels, and the network will then affected by the hard quantization error. Moreover, the clustering is applied based on the clustering result in previous iterations, which accumulates the clustering error during iterations.
On the contrary, we propose a framework that mines the similarity as a soft constraint. By regarding each training image as a different class and training with the softened label distribution, we avoid quantization loss and provide more room for the algorithm.

\section{Proposed Method}
In this paper, we focus on the \textit{unsupervised} re-ID problem. Given a training set of pedestrian images, we aim to learn a feature embedding function for the person images by exploring the image relationship instead of using human annotations. Then, in the evaluation stage, for both query data and gallery data, we use the learned feature embedding function to embed each image into the feature space. The query result is a ranking list of all testing images according to the Euclidean distance between the feature embedding of the query and testing data.

Under the unsupervised setting, the image labels are unknown, so that we regard each image as a different class to initialize a network and gradually mine similarity among unlabeled images as gentle supervision. 
As illustrated in Figure \ref{fig:pipline}, our framework combines three sub-components (shown in three colored rectangles) : (1) A baseline classification network is adopted to classify each image into different classes. The baseline is used as initialization to generate feature representations; (2) The similarity between unlabeled images is explored based on the feature embedding and the auxiliary information to select reliable images for each training data; (3) The target label distribution is softened according to the reliable images, and the network is fine-tuned with the softened labels to pull the selected reliable images together and repel the other images.

\subsection{Baseline: Initialization with Hard Labels} \label{sec: baseline}
Under an unsupervised person re-ID setting, suppose we have a training set $\mathcal{X}=\{x_1, x_2,..., x_N\}$, where each $x_i$ is an unlabeled person image. Our goal is to learn a feature embedding function $\phi(\theta;x_{i})$ from $\mathcal{X}$ without any manual annotation, where parameters of $\phi$ are collectively denoted as $\theta$. Since we do not have ground truth identity label for each image $x_i$, initially we assign each training data $x_i$ by its index, \emph{i.e.,} $\{y_{i}=i~|~1\le i\le N\}$. $y_{i}$ is the initial pseudo label for data $x_i$. In this way, each training image is assumed to fall into an individual class by itself.

Following \cite{xiao2017joint,wu2018unsupervised,lin2019aBottom}, we adopt the classification model with a non-parametric classifier, where a lookup table is used to store the features of all training images. The stored feature of each image is then used as the weight vector of each class.
We formulate the classification objective using the softmax criterion. 
For each image $x$, we normalize its feature $||\bm{v}||=1$ via $\bm{v} = \frac{\phi(\bm{\theta};x)}{||\phi(\bm{\theta};x)||}$. 
Then the probability of an image belongs to the $i$-th class is defined as:
\begin{equation}\label{eq: probability}
p(y_i|x,\bm{V}) = \frac{\exp(\bm{V}_{i}^{\top}\bm{v}/\tau)}{\sum_{j=1}^{N}exp(\bm{V}_j^\mathrm{T}\bm{v}/\tau)},
\end{equation}
where $\bm{V} \in \mathbb{R}^{N \times n_{\phi}}$ is the lookup table that stores the feature of each class, $\bm{V}_{j}$ is the $j$-th column of $\bm{V}$ which indicates the feature of $j$-th class. $N$ is the number of classes, which is the same as the number of training images.
$\tau$ is a temperature parameter \cite{hinton2015distilling} that controls the softness of probability distribution over classes. We set $\tau=0.1$ following \cite{xiao2017joint}.

The loss function is formulated as:
\begin{align}\label{eq: loss}
	\mathcal{L} = - \sum_{j=1}^{N} \log( p(y_{j} | x_i,\bm{V}) t(y_{j}),
\end{align}
where $t(y_j)$ is the conditional empirical distribution over class labels. We set the probability of the distribution to 1 for the ground truth class, and 0 for all other classes. 
The objective Eq. \ref{eq: loss} maximizes the cosine distance between each image feature $\bm{v}_{i}$ and each features in the lookup table $\bm{V}_{j \neq y_{i}}$, while minimizes the cosine distance between each image feature $\bm{v}_{i}$ and the corresponding centroid feature $\bm{V}_{j=y_i}$.

\subsection{Model Learning with Softened Similarity} \label{sec: train}
The initialized baseline network learns to recognize each unlabeled image and obtains an initial discriminative ability. 
By Eq. \ref{eq: probability}, each training sample is learned to push other training images away. 
However, there are images of the same identity, which are supposed to be close in the feature space. Forcing the images of the same person to have obviously different representations will have a negative effect on the network.
Inspired by ECN \cite{zhong2019invariance,zhong2020learning}, we propose to learn a similar representation for images that estimated to be the same identity. 

To find the images of the same identity, we select images with the smallest dissimilarity for each training sample. 
For two images $x_a$, and $x_b$, we define the dissimilarity between the two images as the distance between two images, \ie, $D(x_a, x_b) = d(x_a,x_b)$, where the distance is calculated as the Euclidean distance between the two image features, \ie, $d(x_a,x_b) =\|\phi(\bm{\theta};x_a) - \phi(\bm{\theta};x_b)\|$. 
Then for each training image $x_i$, $k$ images with the smallest dissimilarity are selected as reliable images.
We define a reliable image set $\mathcal{X}_i^{\mathrm{reliable}} = \{ x_i^1, x_i^2,... x_i^k\}$ with label $\mathcal{Y}_i^{\mathrm{reliable}} = \{y_i^1, y_i^2,... y_i^k\}$. Each element $x_i^j$ is estimated to be the same identity as $x_i$, and each class $y_i^j$ is regarded as reliable class. 

Instead of taking reliable images as the same class for training, we propose a softened classification network that learns the similarity among identities in a more smooth way. During training, we want the network could not only predict each image into the ground truth class, but make it acceptable to predict the training image into reliable classes. 
Therefore, we re-assign a nonzero value to the reliable classes in the target label. The target label distribution for data $x_i$ is then written as:
\begin{equation} \label{eq: prob2}
 t(y_j)=\left\{
\begin{array}{lr}
\lambda, & y_j = y_i \\
(1-\lambda)/k, & y_{j} \in \mathcal{Y}_{i}^{\mathrm{reliable}} \\
0, & \mathrm{otherwise}
\end{array}
\right.,
\end{equation}
where $\lambda$ is a hyper-parameter that balances the effect of the ground truth class and the reliable classes. When $\lambda$ is 1, Eq.~\ref{eq: prob2} reduces to the function with only 0,1 options in the baseline network, that the model learns to recognize each image but fails to learn the similarity and consistency among images of the same person. On the other hand, when $\lambda$ is too small, the model may fail to predict the ground truth label. 

Comparing with the baseline network, the images are labeled with soft label distribution (which denotes probabilities) rather than hard 0,1 labels. The labels are no longer the ground truth classes, but probabilities over $k$ possible reliable classes.
By taking reliable classes into account, the confidence of the ground-truth class is reduced, and the confidence of the reliable classes is increased, which guides the network to learn the similarity among images of the identity smoothly. 
With Eq. \ref{eq: loss} and Eq. \ref{eq: prob2}, we define the softened cross-entropy loss as:

\begin{equation}
\begin{aligned}\label{eq: loss3}
	\mathcal{L} = &- \lambda \log( p({y}_{i} | x_i,\bm{V}) \\ 
	&- \frac{1-\lambda}{k}\sum_{j=1}^{k} \log( p(y_i^j | x_i,\bm{V}) ,
\end{aligned}
\end{equation}

The proposed objective not only minimizes the cosine distance between each image feature and the ground truth feature in the lookup table, but minimizes the distance between the features of each image and its reliable images. Meanwhile, the cosine distance between each image feature and the features from other classes is maximized. 

With the softened classification network, we gradually learn a feature that is close to reliable images. 
The learning of the reliable classes is soft and gentle that tries to avoid the negative impact when we involve wrong images in the reliable set. On the other hand, the relatively weak supervision signal makes the model freer and has a higher potential. In this way, we could leverage auxiliary information to help learn a better model. In experiments, we validate that with auxiliary information, the softly learned model performs better than the model learned with hard labels, and will discuss later in Section \ref{sec: compare_buc}.

\subsection{Similarity Estimation with Auxiliary Information} \label{sec: similarity}
As illustrated in Section \ref{sec: train}, for each training sample, $k$ images with the smallest dissimilarity are selected to be reliable. 
To introduce additional priors for constraints, we also think of other resources to help estimate the similarity.

\textbf{Part similarity exploration.}
To assist the similarity measurement between the global feature, we propose to consider the similarity between part features (details) as well. Following~\cite{sun2018beyond}, we extract the CNN feature map and divide it into $p$ horizontal stripes. 
Each partition feature is then average pooled to be a part-level feature embedding. 
We take the average distance of the corresponding parts as the part distance between two images. 
The part distance between two images $x_a$ and $x_b$ is then formulated as:
\begin{equation}
    d_{\mathrm{part}}(x_a,x_b) = \frac{\sum_{i=1}^{p}\|\phi^i(\bm{\theta}, x_a) - \phi^i(\bm{\theta}, x_b) \|} {p},
\end{equation}
where $\phi^i$ is the $i$-th part feature embedding function.

\begin{figure}[!t]
\begin{center}
\includegraphics[width=\linewidth]{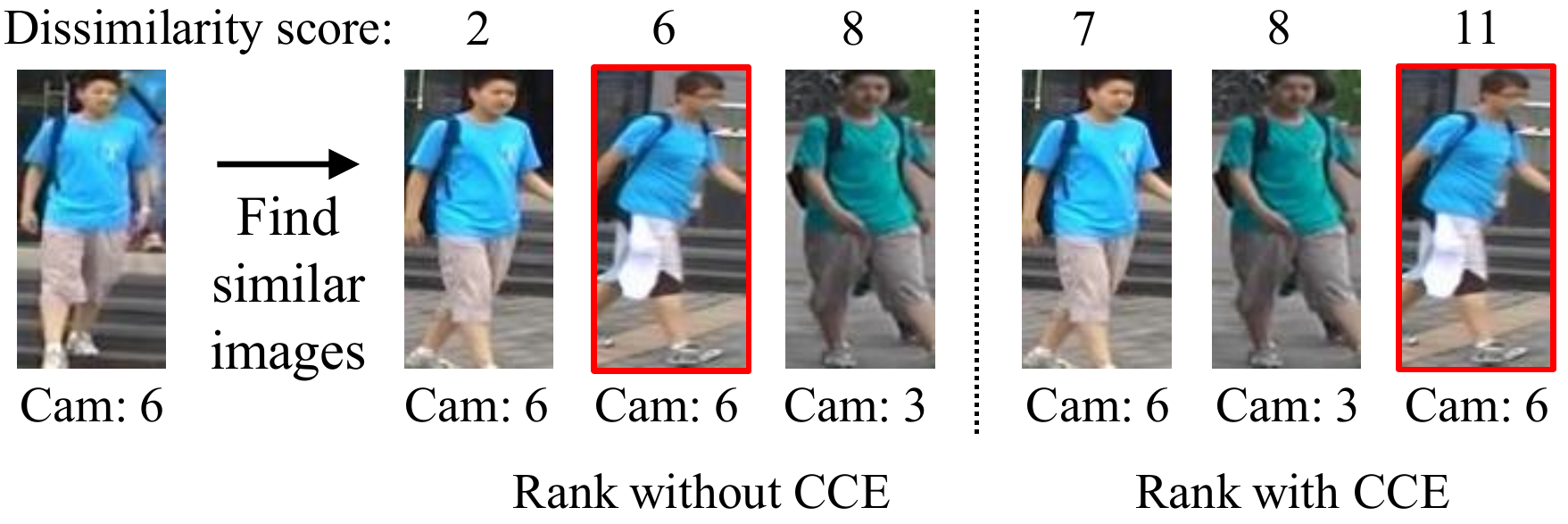}
\end{center}
\caption{Illustration of the cross-camera encouragement term. When calculating the dissimilarity with and without CCE, the chosen reliable images are different. CCE promotes to find the cross-camera ground truth, instead of the hard negative sample. The negative images are shown in red. 
}
\label{fig:CameraPenalty}
\end{figure}
\textbf{The cross-camera encouragement.}
We propose a cross-camera encouragement term (CCE) that added to the dissimilarity to promote images captured by different cameras be viewed as reliable images. 
The intuition of adding CCE is two-fold. First, comparing to inner-camera pairs, the image pairs of different camera ID would teach the network to learn cross-camera information. As a result, the model predicts similar features for a person under different camera views, which benefits the re-ID task.
Second, there are many different pedestrians wearing similar clothes that appear under the same camera. CCE helps to find the cross-camera ground truth, instead of these hard negative samples.
As shown in Figure \ref{fig:CameraPenalty}, without CCE, although the query and the image captured by camera 3 belong to the same person, their dissimilarity is large (8) due to the camera gap. Even a negative example (the one in red) has a smaller distance to the query since they come from the same camera.

Specifically, we denote the camera ID of the training samples as $\mathcal{C} = \{c_1, c_2, ...,c_N\}$. 
The CCE between two images $x_a$ and $x_b$ is formulated as:
\begin{equation}
 \mathrm{CCE}(x_a, x_b)=\left\{
\begin{array}{lr}
\lambda_c,& c_a = c_b \\
0,& c_a \neq c_b 
\end{array}
\right.,
\end{equation}
where $\lambda_c$ is the parameter that controls the strength of cross-camera promotion. 
With the CCE term, the dissimilarity between images with the same camera ID is increased. Thus CCE helps to incorporate more cross camera images in the reliable set, and reduce some inner-camera negative images.

\textbf{Overall dissimilarity.}
Considering the part similarity exploration and the cross-camera encouragement, the overall dissimilarity $D(x_a, x_b)$ between the image $x_a$ and $x_b$ is then formulated as:
\begin{equation}
\begin{aligned}
D(x_a, x_b) = (1-\lambda_p) d(x_a,x_b)&+\lambda_p d_{\mathrm{part}}(x_a,x_b)\\&+ \mathrm{CCE}(x_a, x_b),
\end{aligned}
\end{equation}
where $\lambda_p$ balances the contribution of the global and part similarity. As shown in the green component of Figure \ref{fig:pipline}, the dissimilarity between two images consists of the global distance, the part distance, and the cross-camera encouragement term. By computing the global and the part distance, the similarity of the global appearance and local details are measured, which guarantees the accuracy of reliable image selection. By adding the CCE term, images from different cameras tend to be selected as reliable ones, which enables the network to learn from diverse images. Both of them benefit the discriminative ability of the trained model.

 \begin{table*}[!t]
    \centering
        \begin{tabular}{l|l|cccc|cccc}
            \hline
            \renewcommand{\multirowsetup}{\centering} 
            \multirow{2}{*}{Methods}&\multirow{2}{*}{Setting}&\multicolumn{4}{c|}{Market-1501}&\multicolumn{4}{c}{DukeMTMC-reID} \\
            \cline{3-10}&&rank-1 & rank-5 & rank-10 &mAP&rank-1 & rank-5 & rank-10 &mAP\\
            \hline
          OIM \cite{xiao2017joint}&\textbf{Unsupervised} &38.0&58.0&66.3&14.0&24.5&38.8&46.0&11.3\\
          EUG \cite{wu2018exploit}&OneEx&49.8&66.4&72.7&22.5&45.2&59.2&63.4&24.5\\
          ATNet \cite{liu2019adaptive} &UDA&55.7&73.2&74.9&25.6&45.1&59.5&64.2&24.9\\
          ProLearn \cite{wu2019progressive}&OneEx&55.8 &72.3& 78.4& 26.2& 48.8& 63.4 &68.4 & 28.5\\
          SPGAN \cite{deng2018image}&UDA&58.1&76.0&82.7&26.7&46.9&62.6&68.5&26.4\\
          TJ-AIDL \cite{Wang_2018_CVPR} &UDA&58.2&-&-&26.5&44.3&-&-&23.0\\
		BUC \cite{lin2019aBottom} &{\bf Unsupervised}  &61.0&71.6&76.4&30.6&40.2&52.7&57.4&21.9\\
          HHL \cite{zhong2018generalizing} &UDA&62.2&78.8&84.0&31.4&46.9&61.0&66.7&27.2\\
				\hline
		\textbf{Baseline} &{\bf Unsupervised}  &34.4&54.1&62.3&13.2&16.5&29.9&37.3&7.9\\
		\textbf{Ours (w/o part and CCE)}  &{\bf Unsupervised}  &58.7&70.4&76.3&29.8&31.6&48.3&53.4&17.4\\
		\textbf{Ours (w/o part)} &{\bf Unsupervised}  &68.4&80.8&84.1&35.1&49.2&61.3&65.8&26.4\\
		\textbf{Ours} &{\bf Unsupervised}  & \textbf{71.7}&\textbf{83.8}&\textbf{87.4}&\textbf{37.8}&\textbf{52.5}&\textbf{63.5}&\textbf{68.9}&\textbf{28.6}\\
            \hline
        \end{tabular}
    \caption{
            Comparison with the state-of-the-art methods on the two image-based re-ID datasets, \emph{i.e.}, the Market-1501 dataset and the DukeMTMC-reID dataset. In the column ``Setting'', ``UDA'' denotes the unsupervised domain adaptation methods. ``OneEx'' denotes the methods use the one-example annotation, in which each person in the dataset is annotated with one labeled example. 
        }
        \label{table:image_results}
\end{table*}

 \begin{table*}[!t]
    \centering
        \setlength{\tabcolsep}{3mm}{
        \begin{tabular}{l|l|cccc|cccc}
            \hline
            \renewcommand{\multirowsetup}{\centering} 
             \multirow{2}{*}{Methods}&\multirow{2}{*}{Setting}&\multicolumn{4}{c|}{MARS}&\multicolumn{4}{c}{DukeMTMC-VideoReID} \\
            \cline{3-10}&&rank-1 & rank-5 & rank-10 &mAP&rank-1 & rank-5 & rank-10 &mAP\\
            \hline
            
			OIM \cite{xiao2017joint} & {\bf Unsupervised} & 33.7 & 48.1 & 54.8 & 13.5 & 51.1 &70.5&76.2&43.8\\
            DGM+IDE \cite{ye2017dynamic}      & OneEx      & 36.8 & 54.0 & -    & 16.8 & 42.3 &57.9 &69.3 &33.6\\
			Stepwise \cite{liu2017stepwise}  & OneEx      & 41.2 & 55.5 & -    & 19.6 & 56.2 &70.3 &79.2 &46.7\\
			RACE \cite{eccv18race}           & OneEx      & 43.2 & 57.1 & 62.1 & 24.5 & -&-&-&-\\
			DAL \cite{chen2018deep}          & \textbf{Unsupervised}       & 49.3 & 65.9 & 72.2 & 23.0 & -&-&-&-\\
			BUC \cite{lin2019aBottom}       &{\bf Unsupervised}  &57.9&72.3&75.9&34.7&76.2&88.3&91.0&68.3\\
			EUG \cite{wu2018exploit}         & OneEx      & 62.6 & 74.9 & -    & 42.4&72.7&84.1&-&63.2\\
				\hline
			\textbf{Ours}                &{\bf Unsupervised}  & 	\textbf{62.8}&\textbf{77.2}&\textbf{80.1}&\textbf{43.6}&\textbf{76.4}&\textbf{88.7}&\textbf{91.0}&\textbf{69.3}\\
            \hline
        \end{tabular}}
        \caption{
            Comparison with the state-of-the-art methods on two video-based re-ID datasets, MARS and DukeMTMC-VideoReID. In the column ``Setting'', ``OneEx'' denotes the methods use the one-example annotation, in which each person in the dataset is annotated with one labeled example. 
            ``UDA'' denotes the unsupervised domain adaptation methods.
        }
    \label{table:video_results}
\end{table*}

\begin{figure*}[!t]
\begin{center}
\includegraphics[width=\linewidth]{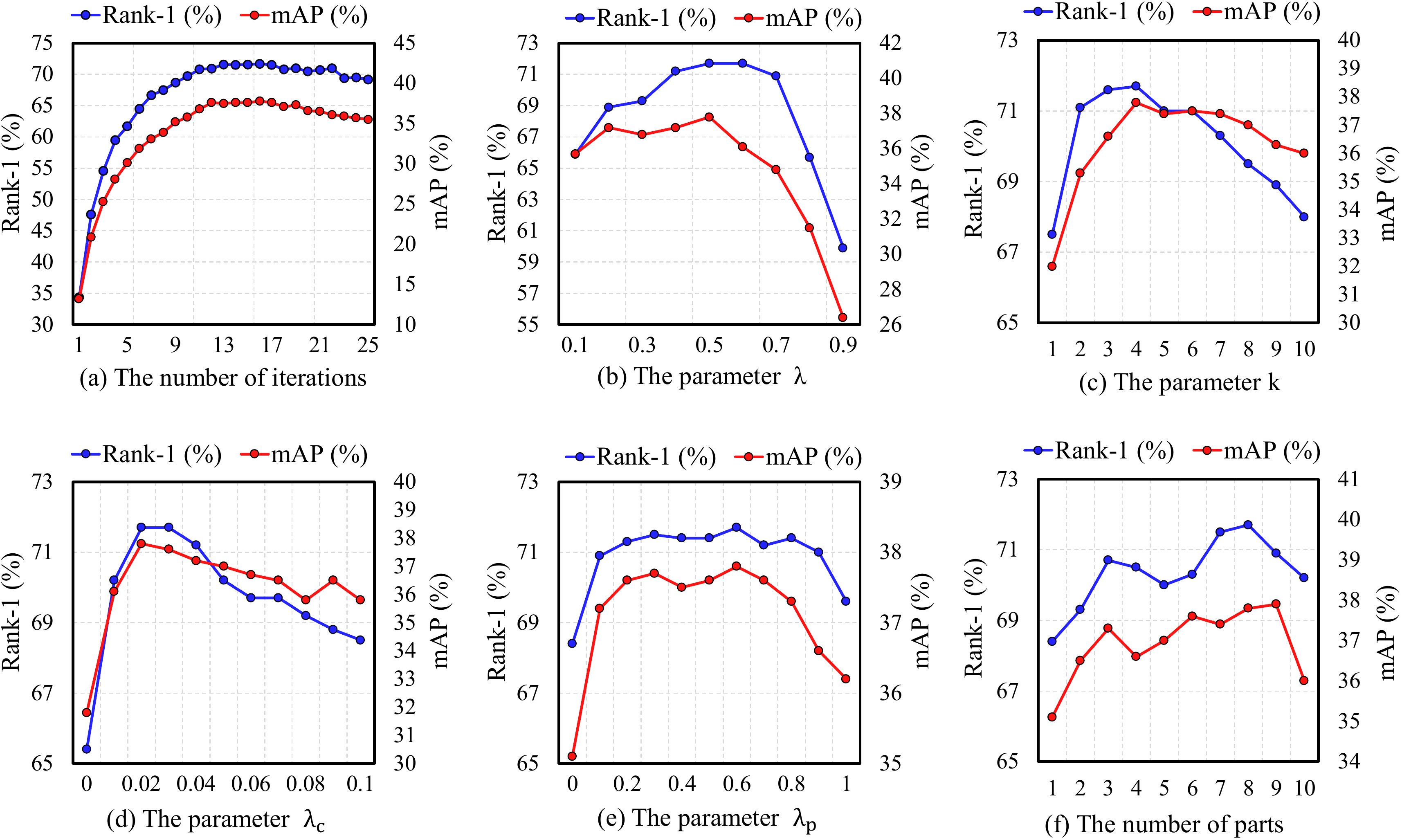}
\end{center}
\caption{Parameter and method analysis on Market-1501. (a) The performance along with iterations. (b) The impact of $\lambda$ for softened classification. (c) The impact of the number of reliable images $k$. (e) The impact of $\lambda_c$ for CCE. (e) The impact of $\lambda_{p}$ for part distance. (f) The impact of the number of parts.
}
\label{fig:parameter2}
\end{figure*}
\section{Experiments}

\subsection{Datasets and Implementation Details}

\textbf{The Market1501 dataset} \cite{zheng2015scalable} is a large-scale dataset captured by 6 cameras for person re-ID. It contains 751 identities for training and 750 identities for testing. The training set, gallery set and query set contain 12936 images, 19732 images and 3368 query images, respectively.  

\textbf{The DukeMTMC-reID dataset} \cite{zheng2017unlabeled} is a subset of the DukeMTMC dataset \cite{ristani2016performance}. It contains 1812 identities captured by 8 cameras. Using the evaluation protocol specified in \cite{zheng2017unlabeled}, we obtain 2228 query images, 16522 training images and 17661 gallery images.  

\textbf{The MARS dataset} \cite{zheng2016mars} is a large-scale video-based dataset for person re-ID. The dataset contains 17503 video tracklets of 1261 identities, where 625 identities are used for training and 636 identities are used for testing. 

\textbf{The DukeMTMC-VideoReID dataset} \cite{wu2018exploit} is a video-based re-ID dataset derived from the DukeMTMC dataset~\cite{ristani2016performance}. It contains 2196 tracklets of 702 identities for training, 2636 tracklets of other 702 identities for testing.

\textbf{Implementation details.}
We adopt ResNet-50 as the CNN backbone and initialize it by the ImageNet \cite{krizhevsky2012imagenet} pre-trained model with the last classification layer removed.
The number of training epochs for the baseline network is set to be 25 for image-based datasets and 30 for video-based, the batch size is set to be 16, the dropout rate is set to be 0.5. The $\lambda$ is set to 0.6. The $\lambda_{p}$ and $\lambda_c$ are set to be 0.5 and 0.02 respectively. The number of parts $p$ is set to be 8.
We use stochastic gradient descent with a momentum of 0.9 to optimize the network. The learning rate is initialized to 0.1 and changed to 0.01 after 15 epochs. For video-based datasets, we take the average feature of all frames within a tracklet to be the tracklet feature.
We implement our method on both PaddlePaddle and PyTorch.
On Market-1501 and DukeMTMC-reID, it takes about 4 hours to finish the training procedure with a GTX 1080TI GPU. On Mars and DukeMTMC-VideoReID, it takes about 12 hours.

\subsection{Comparison with the State-of-the-Arts}
\textbf{Image-based Person Re-identification.} The comparisons with the state-of-the-art algorithms on Market-1501 and DukeMTMC-reID are shown in Table \ref{table:image_results}. 
On Market-1501, under the same setting, we obtain the best performance among the compared methods with \textbf{rank-1 = 71.7\%, mAP = 37.8\%}.
Compared to the state-of-the-art unsupervised method BUC \cite{lin2019aBottom}, we achieve 10.7 points and 7.2 points improvement on rank-1 accuracy and mAP, respectively.
On DukeMTMC-reID, compared to BUC, our method achieves 12.3 and 6.7 points of improvement on rank-1 accuracy and mAP, respectively.
The impressive performance indicates that the softened similarity learning successfully finds images of the same identity and encourages reliable images gathered in the feature space. The proposed CCE helps to learn a discriminative model cross-camera, while the part similarity estimation helps to maintain an accurate reliable image selection. 

\textbf{Video-based Person Re-identification.}
The comparisons with the state-of-the-art algorithms on the two video-based datasets are shown in Table \ref{table:video_results}.  On MARS, we obtain rank-1 = 62.8\%, mAP = 43.6\%. Compared to BUC \cite{lin2019aBottom}, We achieve 4.9 and 8.9 points of improvement in rank-1 accuracy and mAP, respectively. On DukeMTMC-VideoReID, we achieve rank-1 of 76.4\% and mAP of 69.3\%, which beats BUC by 0.2 and 1.0 points, respectively. 
The performance gap between ours and BUC is relatively small on DukeMTMC-VideoReID. We suspect that, the rank-1 accuracy of BUC is 76.2\%, which is quite high under the unsupervised setting, and it is more difficult to make progress upon a high performance. 
Note that without any annotation, we still beat the EUG method in the one-example setting, where each person is annotated a tracklet as labeled data.

\subsection{Diagnostic Studies}\label{sec: ablation}
\textbf{Robustness test.}
Figure \ref{fig:parameter2} (a) illustrates the re-ID performance over iterations. Throughout iterations, the rank-1 accuracy constantly increases from 34.4\% to 71.7\%, which indicates that the model grows robust steadily. After the 16\textsuperscript{th} iteration, the re-ID performance stops to increase and shows a slight decrease. Note that, from the 10\textsuperscript{th} iteration to the 25\textsuperscript{th} iteration, our method always maintains a high re-ID performance, \ie{ with a rank-1 accuracy higher than 69\%}, which demonstrates the robustness of the proposed method.

\textbf{The impact of the hyper-parameter $\lambda$.}
In Eq. \ref{eq: prob2}, the hyper-parameter $\lambda$ controls the degree of softening, which balances the impact of the ground truth class and the selected reliable classes. When $\lambda$ is 0, each training image is learned to be predicted to the reliable classes. When $\lambda$ is 1, each training image will be predicted to its own ground truth class.
We vary $\lambda$ from 0.1 to 0.9 in Figure \ref{fig:parameter2} (b), and observe that, when $\lambda$ increases from 0.1 to 0.6, the re-ID performance continue increasing. When $\lambda$ continually gets larger, we observe a obvious drop on re-ID performance.

\textbf{The impact of the number of reliable images $k$.}
Figure~\ref{fig:parameter2}~(c) shows how the re-ID performance varies with different numbers of reliable images, $k$. We observe that as $k$ increases from 0 to 4, the re-ID performance continues rising, and the performance begins to drop when $k$ gets larger. The reason is that when $k$ is too small, the learned similarity of one identity is not adequate, which makes the model difficult to match images of the same identity. When $k$ is too large, error cases will be involved in the reliable set, which harms the network training when forcing images of different persons to get closer. 

\textbf{The impact of the cross-camera encouragement term.} As shown in Table \ref{table:image_results}, on Market-1501, the results of Ours (w/o part) beat Ours (w/o part and CCE) by 9.7 and 5.3 points on rank-1 and mAP respectively. On DukeMTMC-reID, the improvements on rank-1 and mAP are 17.6 and 9.0, respectively. The impressive improvements demonstrate the effectiveness of the CCE module. Without CCE, images of one identity from different cameras are hard to be selected as reliable images because of the camera variance. 
CCE encourages cross-camera image selection, which enables the model learning from diverse images and getting robust to camera views. 
Besides, we evaluate CCE based on BUC \cite{lin2019aBottom}, and the result is shown in Table \ref{table:compare_buc}. 
We observe that on Market-1501, the improvements of using CCE are 4.9 points for rank-1 and 1.2 points for mAP. This further indicates that CCE is effective and can be easily adopted to other unsupervised methods to achieve better performance.

The parameter $\lambda_{c}$ of CCE balances the effect of appearance and camera diversity. We evaluate different values of $\lambda_{c}$ in Figure \ref{fig:parameter2} (d). As $\lambda_c$ increases from 0 to 0.02, the rank-1 accuracy on Market-1501 increases from 65.4\% to 71.7\%. If we set $\lambda_{c}$ to be greater than 0.02, the too large encouragement term would lead to a negative effect on the performance.

 \begin{table}[!t]
 \footnotesize
    \centering
        \begin{tabular}[c]{l|l|cc|cc}
            \hline
            \renewcommand{\multirowsetup}{\centering} 
    
    \multirow{2}{*}{Dataset}&\multirow{2}{*}{Auxiliary}& \multicolumn{2}{c|}{Ours} &\multicolumn{2}{c}{BUC} \\
    \cline{3-6}&&rank-1 &mAP&rank-1  &mAP\\
    
	\hline
	\multirow{3}{*}{Market-1501}&None  &58.7&29.8&61.0&30.6\\
	\cline{2-6}&CCE   &68.4&35.1& 65.9&31.8\\
	\cline{2-6}&CCE+part &\textbf{71.7}&\textbf{37.8}&69.5&36.2\\
    \hline
    \multirow{3}{*}{DukeMTMC}&None  &31.6&17.4&40.2&21.9\\
	\cline{2-6}&CCE   &49.2&26.4& 48.3&24.4\\
	\cline{2-6}&CCE+part &\textbf{52.5}&\textbf{28.6}&51.5&25.1\\
    \hline
    \end{tabular}
    \caption{
            Comparison with BUC \cite{lin2019aBottom} on Market-1501 and DukeMTMC-reID. The column “Auxiliary” lists the auxiliary information utilized by the method. 
        }
    \label{table:compare_buc}
\end{table}

\textbf{The impact of part similarity.}
As shown in Table \ref{table:image_results}, on Market-1501, the results of Ours beat ``Ours (w/o part)'' by 3.3 and 2.7 points on rank-1 accuracy and mAP, respectively. On DukeMTMC-reID, the improvements on rank-1 and mAP are 3.3 and 2.2, respectively. We also evaluate the part similarity based on BUC \cite{lin2019aBottom}, and the result is shown in Table \ref{table:compare_buc}. We observe that on Market-1501, the improvements of using the part similarity are 3.6 points for rank-1 and 4.4 points for mAP.
This demonstrates that investigating the appearance between pedestrian parts is beneficial for similarity estimation. This idea is also effective on other unsupervised methods and can be easily adopted. 

The parameter $\lambda_p$ balances the effect of the global distance and part distance. We evaluate different values for the parameter $\lambda_p$ in Figure \ref{fig:parameter2} (e). When $\lambda_p=0$, we only adopt the global distance. As $\lambda_p$ increases, retrieval accuracy improves at first. When $\lambda_p=0.5$, we obtain the best performance. After that, the performance begins to drop.

In Figure \ref{fig:parameter2} (f), we vary the number of parts $p$ from 1 to 10. When $p=1$, the part distance is the same as the global distance. When $p=8$, we obtain the best accuracy.

\subsection{Delving into the Softened Similarity Learning and the Hard Label Learning} \label{sec: compare_buc}
We conduct experiments to examine the effectiveness of our method of softened similarity learning and BUC that learn from hard labels. The models learning with and without the auxiliary information are compared. The experimental results are summarized in Table \ref{table:image_results} and Table \ref{table:compare_buc}.

We first observe that both approaches without the auxiliary information yield improvement over the baseline. As shown in Table \ref{table:image_results}, the rank-1 of the baseline and Ours (w/o part and CCE) are 34.4\% and 58.7\%, respectively. The large performance gap demonstrates the effectiveness of softened similarity learning. 
Second, as shown in Table \ref{table:compare_buc}, without any auxiliary information, BUC achieves better performance than ours on both datasets. We think the reason is that our network is trained with softened labels, which avoids pushing images of the same identity away, but it is also has a relatively small strength to force images of different identities separate. Nevertheless, from Table \ref{table:compare_buc}, we find that when we adopt CCE or part similarity into the two approaches, our method exceeds BUC on both datasets. This indicates that given a better similarity estimation, the softened similarity learning has a higher potential to learn better embeddings. We suspect that when learning with hard one-hot labels, the model is forced to fit the noise labels, which limits its accuracy. In contrast, our method hardly affected by the inaccurate similarity estimation and thus has more room to learn and improve.
\section{Conclusions}

In this paper, we investigate the problem of unsupervised re-ID. Following the pipeline of iterative person recognition and feature update, we propose not to assign each sample with a hard label so as to avoid quantization loss as well as provide more room for the learning algorithm. To introduce additional priors for constraints, we introduce several auxiliary information, including a camera-based term which is easy to obtain yet useful for distance amendment. Experiments on both image-based and video-based re-ID tasks validate the effectiveness of our approach.

This work puts forward a point that classification may not be the optimal supervision, in particular, for unsupervised re-ID. This reminds us of the difference between classification-based and metric-learning-based methods in supervised re-ID. The potential connections between them remain uncovered, which we will investigate in the future research.

{\bf Acknowledgments.} This work is supported by National Nature Science Foundation of China (61931008, 61671196, 61701149, 61801157, 61971268, 61901145, 61901150, 61972123), National Natural Science Major Foundation of Research Instrumentation of PR China under Grants 61427808
, Zhejiang Province Nature Science Foundation of China (LR17F030006, Q19F010030), 111 Project, No. D17019.

{\small
\bibliographystyle{ieee_fullname}
\bibliography{egbib}
}

\end{document}